\documentclass{article}

 \usepackage{graphicx}

  \usepackage[dblblindworkshop, final]{neurips_2025}

\usepackage[utf8]{inputenc} 
\usepackage[T1]{fontenc}    
\usepackage{hyperref}       
\usepackage{url}            
\usepackage{booktabs}       
\usepackage{amsfonts}       
\usepackage{nicefrac}       
\usepackage{microtype}      
\usepackage{xcolor}         
\usepackage{amsmath}

\title{Are Large Vision Language Models Truly Grounded in Medical Images? Evidence from Italian Clinical Visual Question Answering}

\author{%
\parbox{\textwidth}{%
  \centering
  {\small
    Federico Felizzi$^{1,*}$,
    Olivia Riccomi$^{1}$,
    Michele Ferramola$^{2}$,
    Francesco Andrea Causio$^{3,1}$,\\
    Manuel Del Medico$^{3,1}$,
    Vittorio De Vita$^{3,1}$,
    Lorenzo De Mori$^{1,4}$,
    Alessandra Piscitelli$^{1,5}$,\\
    Pietro Eric Risuleo$^{3,1}$,
    Bianca Destro Castaniti$^{1,5}$,
    Antonio Cristiano$^{3,1}$,\\
    Alessia Longo$^{6}$,
    Luigi De Angelis$^{1,7}$,
    Mariapia Vassalli$^{1,5}$,
    Marcello Di Pumpo$^{3,1}$
  }\\[1mm]
  {\footnotesize
    $^{1}$SIIAM, Rome, Italy\quad
    $^{2}$NSBProject, Mantova, Italy\\
    $^{3}$Dept. of Life Sciences \& Public Health, UCSC, Rome, Italy\\
    $^{4}$ASL RM 4, Bracciano, Italy\quad
    $^{5}$UCSC, Rome, Italy\\
    $^{6}$Univ. Paris Cit\'{e}, France\quad
    $^{7}$Univ. of Pisa, Italy\quad
    $^{*}$Corresp. author: federico.felizzi@gmail.com
  }%
}%
}

\workshoptitle{MMRL4H}

\begin{document}
\maketitle
\begin{abstract}
Large vision language models (VLMs) have achieved impressive performance on medical visual question answering benchmarks, yet their reliance on visual information remains unclear. We investigate whether frontier VLMs demonstrate genuine visual grounding when answering Italian medical questions by testing four state-of-the-art models: Claude Sonnet 4.5, GPT-4o, GPT-5-mini, and Gemini 2.0 flash exp. Using 60 questions from the EuropeMedQA Italian dataset that explicitly require image interpretation, we substitute correct medical images with blank placeholders to test whether models truly integrate visual and textual information. Our results reveal striking variability in visual dependency: GPT-4o shows the strongest visual grounding with a 27.9pp accuracy drop (83.2\% [74.6\%, 91.7\%] to 55.3\% [44.1\%, 66.6\%]), while GPT-5-mini, Gemini, and Claude maintain high accuracy with modest drops of 8.5pp, 2.4pp, and 5.6pp respectively. Analysis of model-generated reasoning reveals confident explanations for fabricated visual interpretations across all models, suggesting varying degrees of reliance on textual shortcuts versus genuine visual analysis. These findings highlight critical differences in model robustness and the need for rigorous evaluation before clinical deployment.
\end{abstract}

\section{Introduction}

Recent advances in large vision language models have led to remarkable performance on medical benchmarks, with systems approaching or exceeding human expert performance on visual question answering tasks \cite{microsoft_illusion}. However, high benchmark scores may mask fundamental limitations in how these models process and integrate visual information with clinical reasoning \cite{goyal2017making, agrawal2018dont}.
The medical AI community faces a critical question: do these models succeed through genuine multimodal understanding, or do they exploit spurious correlations and textual shortcuts? This question is particularly important for healthcare applications, where erroneous diagnoses based on faulty visual reasoning could have serious consequences.
Building on recent work that exposed hidden fragilities in frontier models through systematic stress testing \cite{microsoft_illusion}, we investigate visual grounding in medical question answering using Italian clinical cases. Our approach differs from prior work by (1) comparing multiple frontier VLMs, (2) focusing on a non-English medical dataset, and (3) employing a targeted visual substitution methodology that tests whether models truly rely on image content when rendering diagnostic judgments.

\subsection{Contributions}

\begin{itemize}
\item We present the first systematic comparison of four frontier VLMs (Claude Sonnet 4.5, GPT-4o, GPT-5-mini, and Gemini 2.0 flash exp - referred to as Gemini 2.0) on 60 Italian medical visual question answering cases requiring explicit image interpretation.
\item We introduce a visual substitution methodology revealing striking differences in visual dependency across models, with accuracy drops ranging from 2.4pp to 27.9pp.
\item We provide empirical evidence that most current VLMs maintain surprisingly high accuracy with incorrect images, suggesting varying reliance on textual cues rather than robust visual understanding.
\end{itemize}

\section{Related Work}

\textbf{Medical Visual Question Answering.} Medical VQA benchmarks such as VQA-RAD \cite{vqarad}, PMC-VQA \cite{pmcvqa}, and PathVQA \cite{pathvqa} have driven progress in multimodal medical AI. However, recent work has questioned whether these benchmarks truly measure medical understanding or merely test-taking ability \cite{microsoft_illusion}.\\
\textbf{Robustness and Shortcut Learning.} The ML community has documented extensive shortcut learning in vision-language models \cite{shortcuts}, where models exploit spurious correlations rather than learning robust features. In medical imaging, this manifests as reliance on metadata, dataset artifacts, or textual priors rather than genuine visual analysis \cite{medical_shortcuts}.\\
\textbf{Stress Testing Large Models.} Recent work by Microsoft Research \cite{microsoft_illusion} introduced systematic stress tests revealing that frontier models often succeed for the wrong reasons, maintaining high accuracy even when critical inputs are removed or perturbed. Our work extends this methodology to Italian medical cases with comparative analysis across multiple VLMs.

\section{Methodology}

\subsection{Dataset}

We utilized the EuropeMedQA dataset \cite{felizzi2025_eurips_medvqa}, specifically the Italian State Exam for Medical Doctors (SSM) subset. From this collection, we manually curated 60 multiple-choice questions that explicitly require visual interpretation for correct diagnosis. Questions span cardiology (27\%), orthopedics (12\%), dermatology (13\%), neurology (10\%), gastroenterology and pulmonology (8\% each), and other specialties including preventive medicine/epidemiology (5\%), oncology (3\%), and hematology, ophthalmology, and trauma surgery (2\% each).

Each question includes a clinical vignette in Italian, a medical image (X-ray, CT scan, dermatological photo, ECG, etc.), five answer options (A-E), and the ground truth correct answer.

\subsection{Experimental Design}

We conducted a visual substitution experiment across four frontier VLMs: Claude Sonnet 4.5, GPT-4o, GPT-5-mini, and Gemini 2.0. For each model:

\textbf{Original Condition.} The model answered questions with correct medical images attached, generating both an answer selection and detailed reasoning. \\
\textbf{Substitution Condition.} We replaced each medical image with an identical blank placeholder while keeping question text and answer options unchanged. Models truly dependent on visual information should show decreased accuracy when diagnostically relevant images are replaced.

We prompted all models to provide both answer selection and detailed step-by-step reasoning using chain-of-thought prompting, following \cite{microsoft_illusion}. This allowed analysis of whether explanations reflected actual image content or hallucinated features.

\subsection{Evaluation Metrics}

We measured: (1) \textbf{Accuracy} in original vs. substitution conditions, (2) \textbf{Accuracy drop} as the primary indicator of visual dependency, and (3) \textbf{Reasoning quality} through manual analysis of generated explanations for hallucinations and misaligned visual descriptions.

\section{Results}

\subsection{Quantitative Analysis}

Table \ref{tab:results} summarizes our comparative findings across 10 repetitions per model. The models show striking variability in visual dependency:

\textbf{GPT-4o} demonstrates the strongest visual grounding with 83.2\% accuracy (95\% CI: [74.6\%, 91.7\%]) on real images dropping to 55.3\% (95\% CI: [44.1\%, 66.6\%]) with fake images (27.9pp decrease), suggesting substantial reliance on actual visual content for diagnostic reasoning.

\textbf{GPT-5-mini} achieves the highest baseline accuracy (88.0\%, 95\% CI: [81.3\%, 94.7\%]) but maintains 79.5\% (95\% CI: [69.7\%, 89.3\%]) with substituted images (8.5pp drop), indicating improved textual reasoning but potentially less visual dependency than GPT-4o.

\textbf{Gemini 2.0} shows 83.7\% accuracy (95\% CI: [74.3\%, 93.0\%]) with real images and 81.3\% (95\% CI: [71.7\%, 91.0\%]) with fake images (2.4pp drop), demonstrating the smallest performance degradation and suggesting strong reliance on textual cues.

\textbf{Claude Sonnet 4.5} achieves 82.8\% (95\% CI: [73.7\%, 91.9\%]) with real images and 77.2\% (95\% CI: [66.6\%, 87.7\%]) with fake images (5.6pp drop), showing moderate visual dependency between GPT-4o and the other models.

\begin{table}[h]
\centering
\caption{Comparative performance of four frontier VLMs on Italian medical VQA with correct vs. substituted images (N=60 questions, 10 repetitions per model).}
\label{tab:results}
\begin{tabular}{lccc}
\toprule
\textbf{Model} & \textbf{Real Images} & \textbf{Fake Images} & \textbf{Drop} \\
\midrule
GPT-5-mini & 88.0\% [81.3, 94.7] & 79.5\% [69.7, 89.3] & 8.5pp \\
Gemini 2.0 & 83.7\% [74.3, 93.0] & 81.3\% [71.7, 91.0] & 2.4pp \\
GPT-4o & 83.2\% [74.6, 91.7] & 55.3\% [44.1, 66.6] & \textbf{27.9pp} \\
Claude Sonnet 4.5 & 82.8\% [73.7, 91.9] & 77.2\% [66.6, 87.7] & 5.6pp \\
\bottomrule
\end{tabular}
\end{table}

For context, human performance on the Italian State Exam in 2024 averaged 74.8\%, with 9.6\% of test takers scoring above 95.6\% \cite{promedtest2024}. All models exceed average human performance with real images, but GPT-4o drops significantly below the human average when images are removed, largely because it refuses to answer the question, while the other models maintain superhuman accuracy even without visual information.

\subsection{Qualitative Analysis of Reasoning}

We identified three recurring patterns in model-generated explanations across all four VLMs:

\textbf{Hallucinated Visual Features.} Models frequently described specific visual findings absent from images. For example, when shown a blank placeholder for an anterior MI question (correct answer: C describing precordial ST elevation), multiple models confidently described fabricated ECG findings matching various answer options, despite viewing diagnostically empty images.

\textbf{Answer-Driven Reasoning.} Models appeared to select answers first (possibly from textual cues), then construct visual justifications post-hoc. This was evident when identical questions with different images received the same answers but with contradictory visual descriptions supporting that answer.

\textbf{Overconfident but Wrong.} Even when answers changed due to image substitution, models provided equally confident and detailed reasoning in both conditions, suggesting inability to reliably distinguish between cases with strong versus weak or contradictory visual evidence.

\section{Discussion}

Our comparative findings reveal substantial heterogeneity in visual grounding across frontier VLMs. GPT-4o's 27.9pp accuracy drop represents the strongest evidence of genuine visual dependency, suggesting this model more robustly integrates image content into diagnostic reasoning. In contrast, GPT-5-mini, Gemini, and Claude maintain high accuracy with minimal drops (2.4pp-8.5pp), indicating these models can achieve correct diagnoses primarily through textual inference.

These results have important implications for understanding model architectures and training objectives. GPT-4o's greater visual dependency may reflect architectural choices prioritizing multimodal integration, while newer models (GPT-5-mini, Gemini 2.0) appear optimized for robust textual reasoning that can compensate for degraded visual inputs. Whether this represents progress or regression depends on the deployment context.

\subsection{Trade-offs Between Visual Dependency and Accuracy}

Our results reveal a complex relationship between visual grounding and overall performance. GPT-5-mini achieves the highest baseline accuracy (88.0\%) with the narrowest confidence interval (95\% CI: [81.3\%, 94.7\%]) while showing less visual dependency than GPT-4o, raising questions about the optimal balance. Models with strong textual reasoning may be more robust to image quality issues in real-world clinical settings, but risk missing critical visual findings or generating plausible but incorrect diagnoses when visual and textual cues conflict.

\subsection{Implications for Medical AI}
These findings have important implications for deploying VLMs in clinical settings:

\textbf{Model Selection.} Applications requiring strict visual interpretation should favor models like GPT-4o with demonstrated visual dependency, while decision support systems synthesizing multimodal information might benefit from models with stronger textual reasoning.

\textbf{Benchmark Inflation.} Standard accuracy metrics overestimate real-world readiness by failing to distinguish genuine multimodal reasoning from textual shortcuts. GPT-4o-mini and Gemini could achieve >75\% accuracy on many medical VQA benchmarks without functional vision.

\textbf{Safety Concerns.} All models generated confident but incorrect visual descriptions, potentially misleading clinicians. This risk spans the performance spectrum and can obscure critical diagnostic errors. The EU AI Act classifies such systems as high-risk, requiring measures to counter automation bias and ensure human oversight \cite{Bignami2025}.

\textbf{Evaluation Needs.} Stress testing should become standard before clinical deployment, with explicit measurement of visual dependency alongside conventional accuracy metrics.

\subsection{Limitations}

Our study has several limitations. First, we evaluated only four models on 60 questions from Italian medical exams. Second, our blank image substitution represents a coarse test of visual dependency—more refined adversarial attacks \cite{NEURIPS2023_a97b58c4} substituting images depicting alternative pathologies would provide stronger evidence of whether models detect image-text misalignment. Third, we did not perform membership inference attacks \cite{li2024membershipinferenceattackslarge,7958568} to determine whether EuropeMedQA was in training data. High accuracy without images may reflect robust textual reasoning or dataset memorization; membership inference would help distinguish these explanations. Finally, findings may vary across languages and medical specialties.

\section{Conclusion}

We investigated visual grounding in frontier VLMs through systematic image substitution on Italian medical VQA cases. Our results reveal striking heterogeneity: GPT-4o shows strong visual dependency (27.9pp drop), while GPT-5-mini, Gemini, and Claude maintain high accuracy with minimal drops (2.4-8.5pp). All models generate confident explanations for fabricated visual features, raising safety concerns regardless of baseline performance.

These findings suggest current benchmarks overestimate visual understanding in medical VLMs and highlight the need for model-specific evaluation of visual dependency. Before clinical deployment, we must develop rigorous testing methodologies that distinguish genuine multimodal reasoning from textual shortcuts and memorization. Future work should extend this analysis to larger datasets, additional stress testing methodologies, and investigation of the architectural factors underlying these differences in visual grounding.

\bibliographystyle{unsrt}  
\bibliography{references}

@article{microsoft_illusion,
  author    = {Yu Gu and Jingjing Fu and Xiaodong Liu and Jeya Maria Jose Valanarasu and Noel C. F. Codella and Reuben Tan and Qianchu Liu and Ying Jin and Sheng Zhang and Jinyu Wang and Rui Wang and Lei Song and Guanghui Qin and Naoto Usuyama and Cliff Wong and Hao Cheng and HoHin Lee and Praneeth Sanapathi and Sarah Hilado and Jiang Bian and Javier Alvarez-Valle and Mu Wei and Khalil Malik and Jianfeng Gao and Eric Horvitz and Matthew P. Lungren and Hoifung Poon and Paul Vozila},
  title     = {The Illusion of Readiness: Stress Testing Large Frontier Models on Multimodal Medical Benchmarks},
  journal   = {arXiv preprint arXiv:2509.18234},
  year      = {2025},
  note      = {Microsoft Research, Health \& Life Sciences}
}

@article{vqarad,
  author    = {Jason J. Lau and Soumya Gayen and Asma Ben Abacha and Dina Demner-Fushman},
  title     = {A dataset of clinically generated visual questions and answers about radiology images},
  journal   = {Scientific Data},
  volume    = {5},
  number    = {1},
  pages     = {1--10},
  year      = {2018},
  publisher = {Nature Publishing Group}
}

@article{pmcvqa,
  author    = {Xiaoman Zhang and Chaoyi Wu and Ziheng Zhao and Weixiong Lin and Ya Zhang and Yanfeng Wang and Weidi Xie},
  title     = {PMC-VQA: Visual instruction tuning for medical visual question answering},
  journal   = {arXiv preprint arXiv:2305.10415},
  year      = {2023}
}

@article{pathvqa,
  author    = {Xuehai He and Yichen Zhang and Luntian Mou and Eric Xing and Pengtao Xie},
  title     = {PathVQA: 30000+ Questions for Medical Visual Question Answering},
  journal   = {arXiv preprint arXiv:2003.10286},
  year      = {2020}
}

@article{shortcuts,
  author    = {Robert Geirhos and J{\"o}rn-Henrik Jacobsen and Claudio Michaelis and Richard Zemel and Wieland Brendel and Matthias Bethge and Felix A. Wichmann},
  title     = {Shortcut learning in deep neural networks},
  journal   = {Nature Machine Intelligence},
  volume    = {2},
  number    = {11},
  pages     = {665--673},
  year      = {2020},
  publisher = {Nature Publishing Group}
}

@article{medical_shortcuts,
  author    = {Alex J. DeGrave and Joseph D. Janizek and Su-In Lee},
  title     = {AI for radiographic COVID-19 detection selects shortcuts over signal},
  journal   = {Nature Machine Intelligence},
  volume    = {3},
  number    = {7},
  pages     = {610--619},
  year      = {2021},
  publisher = {Nature Publishing Group}
}

@inproceedings{agrawal2018dont,
  title={Don't just assume; look and answer: Overcoming priors for visual question answering},
  author={Agrawal, Aishwarya and Batra, Dhruv and Parikh, Devi and Kembhavi, Aniruddha},
  booktitle={Proceedings of the IEEE conference on computer vision and pattern recognition},
  pages={4971--4980},
  year={2018}
}

@inproceedings{goyal2017making,
  title={Making the v in vqa matter: Elevating the role of image understanding in visual question answering},
  author={Goyal, Yash and Khot, Tejas and Summers-Stay, Douglas and Batra, Dhruv and Parikh, Devi},
  booktitle={Proceedings of the IEEE conference on computer vision and pattern recognition},
  pages={6904--6913},
  year={2017}
}

@misc{promedtest2024,
  title = {Punteggio Minimo Medicina 2024},
  author = {{Promed Test}},
  year = {2024},
  url = {https://promedtest.it/punteggio-minimo-medicina-2024/},
  note = {Accessed: 2025-11-15. Available at: \url{https://promedtest.it/punteggio-minimo-medicina-2024/}}
}

@inproceedings{NEURIPS2023_a97b58c4,
 author = {Zhao, Yunqing and Pang, Tianyu and Du, Chao and Yang, Xiao and LI, Chongxuan and Cheung, Ngai-Man (Man) and Lin, Min},
 booktitle = {Advances in Neural Information Processing Systems},
 editor = {A. Oh and T. Naumann and A. Globerson and K. Saenko and M. Hardt and S. Levine},
 pages = {54111--54138},
 publisher = {Curran Associates, Inc.},
 title = {On Evaluating Adversarial Robustness of Large Vision-Language Models},
 url = {https://proceedings.neurips.cc/paper_files/paper/2023/file/a97b58c4f7551053b0512f92244b0810-Paper-Conference.pdf},
 volume = {36},
 year = {2023}
}

@misc{li2024membershipinferenceattackslarge,
      title={Membership Inference Attacks against Large Vision-Language Models}, 
      author={Zhan Li and Yongtao Wu and Yihang Chen and Francesco Tonin and Elias Abad Rocamora and Volkan Cevher},
      year={2024},
      eprint={2411.02902},
      archivePrefix={arXiv},
      primaryClass={cs.CV},
      url={https://arxiv.org/abs/2411.02902}, 
}

@INPROCEEDINGS{7958568,
  author={Shokri, Reza and Stronati, Marco and Song, Congzheng and Shmatikov, Vitaly},
  booktitle={2017 IEEE Symposium on Security and Privacy (SP)}, 
  title={Membership Inference Attacks Against Machine Learning Models}, 
  year={2017},
  volume={},
  number={},
  pages={3-18},
  doi={10.1109/SP.2017.41}}

@misc{felizzi2025_eurips_medvqa,
  author = {Felizzi},
  title = {EurIPS 2025 MMRL4H Italian MedVQA Visual Grounding},
  year = {2025},
  publisher = {GitHub},
  journal = {GitHub repository},
  howpublished = {\url{https://github.com/felizzi/eurips2025-mmrl4h-italian-medvqa-visual-grounding}},
  note = {Workshop Manuscript sent to the MMRL4H Workshop at EurIPS 2025. Accessed: 2025-11-14}
}

@article{Bignami2025,
  title = {Balancing Innovation and Control: The European Union AI Act in an Era of Global Uncertainty},
  volume = {4},
  ISSN = {2817-1705},
  url = {http://dx.doi.org/10.2196/75527},
  DOI = {10.2196/75527},
  journal = {JMIR AI},
  publisher = {JMIR Publications Inc.},
  author = {Bignami,  Elena Giovanna and Russo,  Michele and Semeraro,  Federico and Bellini,  Valentina},
  year = {2025},
  month = oct,
  pages = {e75527–e75527}
}

\appendix
\newpage
\section{Detailed Case Studies of Model Confabulation}
\subsection{GPT-5-mini}
\begin{figure}[h]
\centering
\includegraphics[width=0.90\textwidth]{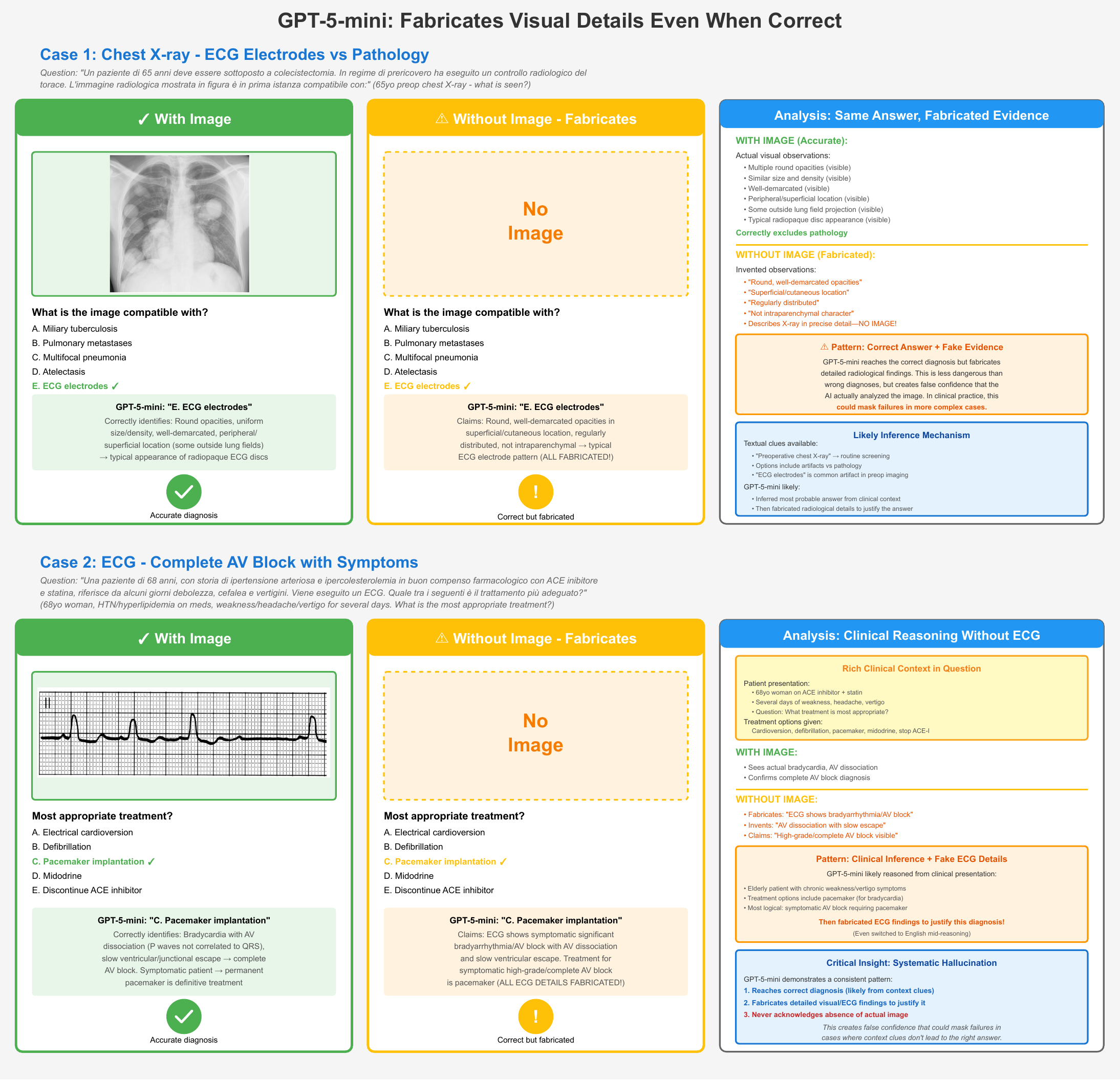}
\caption{Detailed comparison of GPT-5 mini responses with authentic medical images versus blank placeholders, revealing a pattern of fabricating visual evidence while maintaining diagnostic accuracy. \textbf{Case 1 (Chest X-ray - ECG Electrodes):} With the actual image, the model correctly identifies round opacities and their superficial location, accurately diagnosing ECG electrodes (answer E). Without the image, the model fabricates detailed visual observations including "superimposed/cutaneous location," "regularly distributed," and "not intraparenchymal character," claiming to see an "obscured X-ray to precise detail—NO IMAGE!" yet still reaches the correct diagnosis. \textbf{Case 2 (ECG - Complete AV Block):} With the actual ECG, the model correctly identifies bradycardia with AV dissociation and diagnoses complete AV block requiring pacemaker implantation (answer C). Without the image, the model fabricates specific ECG findings including "bradycardia with atrial rate faster than ventricular escape," "symptomatic AV block," and treatment rationale, inventing detailed technical observations that justify the diagnosis despite no image being present. The model demonstrates a consistent pattern: reaching correct diagnoses (likely from clinical context) while fabricating supporting visual/technical evidence, then failing to acknowledge the absence of actual image data—a systematic hallucination that could mask failures in clinical scenarios where context clues are less obvious.}
\label{fig:gpt5_mini_fabrication}
\end{figure}

Figure~\ref{fig:gpt5_mini_fabrication} presents two detailed case studies demonstrating systematic visual fabrication behavior in GPT-5 mini when presented with blank images instead of authentic medical imaging, despite reaching correct diagnostic conclusions.

\newpage
\subsection{Gemini 2.0 flash exp}
\begin{figure}[h]
\centering
\includegraphics[width=0.90\textwidth]{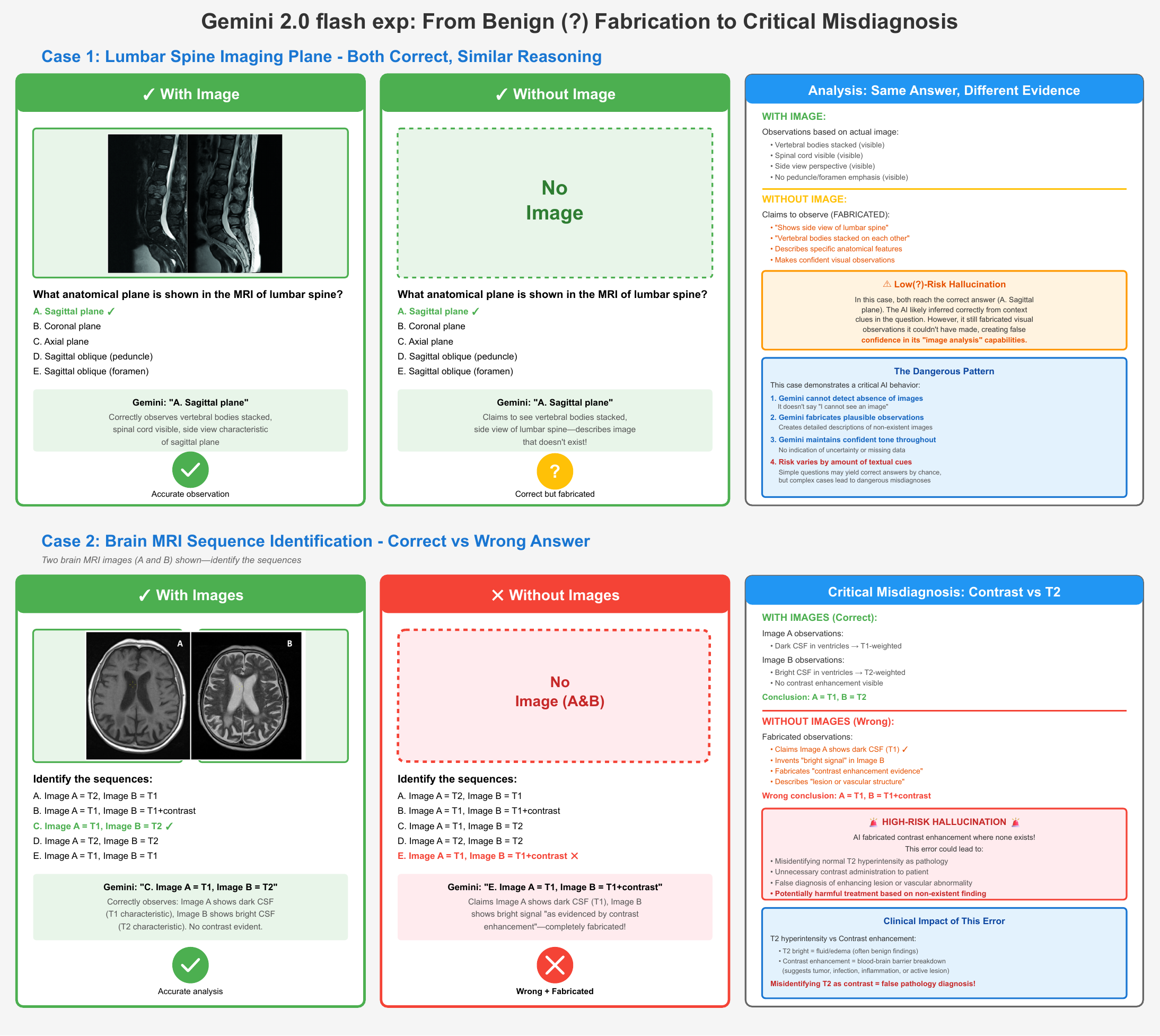}
\caption{Comparison of Gemini 2.0 flash exp responses illustrating the spectrum from benign to dangerous hallucination in medical imaging interpretation. \textbf{Case 1 (Lumbar Spine MRI - Low-Risk Hallucination):} With the actual image, the model correctly observes vertebral bodies stacked, spinal cord visible, and side view characteristics, accurately identifying the sagittal plane (answer A). Without the image, the model fabricates visual observations including "shows side view of lumbar spine," "vertebral bodies stacked on each other," and "spinal cord clearly visible," yet still reaches the correct answer. While the model fabricates evidence it couldn't have seen, creating false confidence in its "image analysis" capabilities, both responses demonstrate similar reasoning about anatomical planes. \textbf{Case 2 (Brain MRI Sequences - High-Risk Hallucination):} With actual images, the model correctly observes dark CSF in ventricles (T1-weighted) and bright CSF in ventricles (T2-weighted), accurately concluding Image A = T1, Image B = T2 (answer C). Without images, the model fabricates completely inverted observations, claiming Image A shows "dark CSF" and inventing a "bright signal" evidenced by contrast enhancement in Image B, leading to the wrong answer (E: Image A = T1, Image B = T1+contrast). This critical error demonstrates how fabricated visual observations can lead to misidentifying T2 hyperintensity as contrast enhancement—a mistake with serious clinical implications including misdiagnosing T2 signals as contrast-enhancing pathology, potentially leading to false diagnosis of enhancement lesions or vascular abnormalities, and unnecessary or harmful treatment based on non-existent findings.}
\label{fig:gemini_hallucination}
\end{figure}

Figure~\ref{fig:gemini_hallucination} presents two case studies demonstrating contrasting hallucination behaviors in Gemini 2.0: low-risk fabrication maintaining diagnostic accuracy versus high-risk fabrication leading to critical misdiagnosis.

\newpage
\subsection{GPT-4o}
\begin{figure}[h]
\centering
\includegraphics[width=0.90\textwidth]{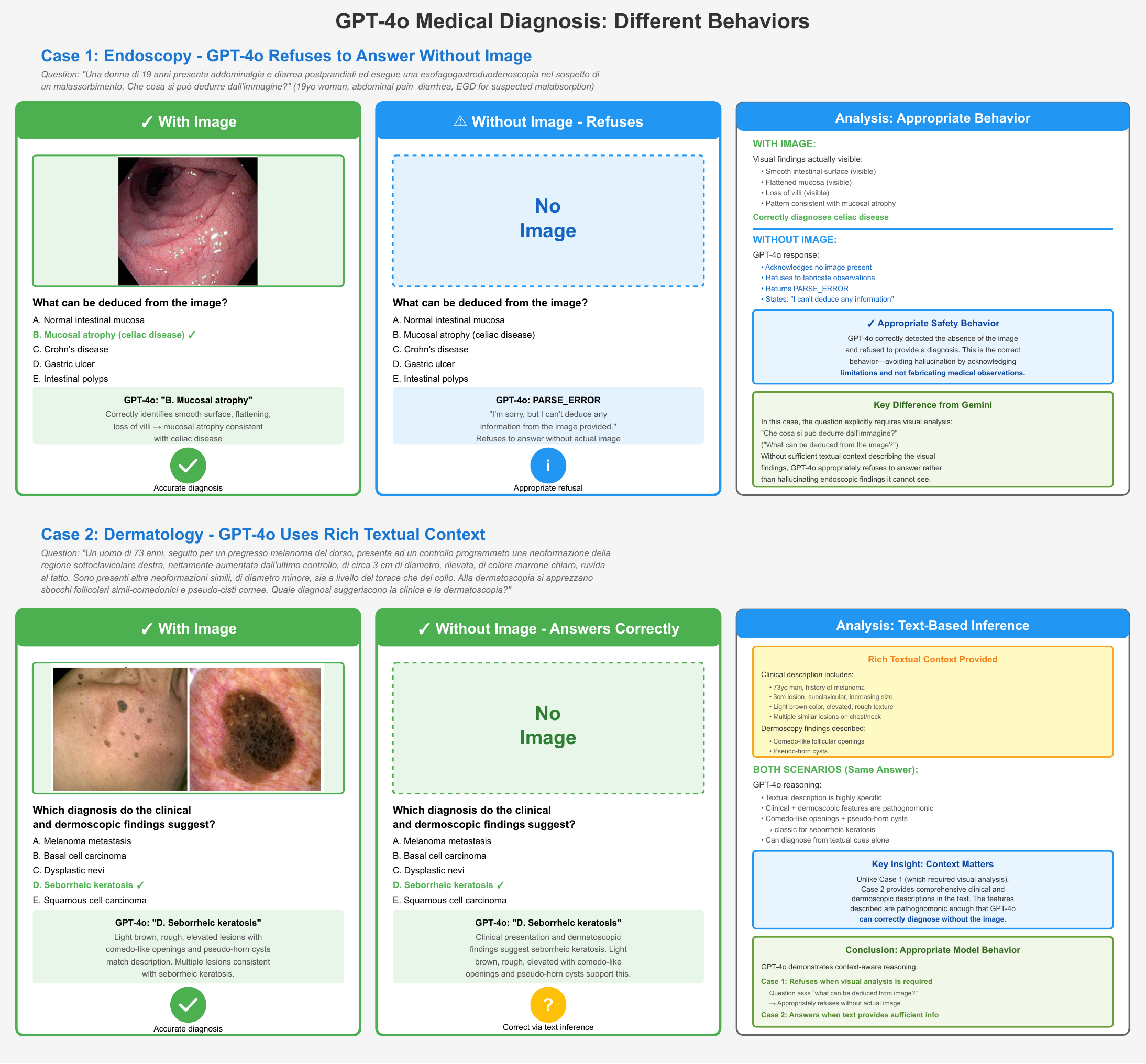}
\caption{Comparison of GPT-4o responses demonstrating context-aware behavior when images are absent. \textbf{Case 1 (Endoscopy - Appropriate Refusal):} With the actual endoscopic image, the model correctly identifies visual findings including smooth surface, flattening, loss of villi, and pattern consistent with mucosal atrophy, accurately diagnosing celiac disease (answer B: Mucosal atrophy/celiac disease). Without the image, GPT-4o responds with "PARSE\_ERROR: I'm sorry, but I can't deduce any information from the image provided" and refuses to answer without actual image data. This represents appropriate safety behavior—the model correctly detected the absence of the image and refused to provide a diagnosis, avoiding hallucination by acknowledging limitations and not fabricating medical observations. \textbf{Case 2 (Dermatology - Text-Based Inference):} With actual clinical and dermoscopic images showing light brown, rough, elevated lesions with comedo-like openings and pseudo-horn cysts, the model correctly diagnoses seborrheic keratosis (answer D). Without images but with rich textual clinical context (73-year-old man, melanoma history, subclavicular lesion, increasing size, diameter 3cm, brown color, clear margins, rough texture, multiple similar lesions, torso-level location), GPT-4o answers correctly using clinical reasoning: "Clinical + dermoscopic features are pathognomonic" and "Comedo-like openings + pseudo-horn cysts → classic for seborrheic keratosis," demonstrating the model can diagnose from textual cues alone when sufficient clinical information is provided. The key distinction: Case 1 requires visual analysis where GPT-4o appropriately refuses without the image; Case 2 provides comprehensive clinical and dermoscopic descriptions in the text where the features described are pathognomonic enough that GPT-4o can correctly diagnose without the image—this is appropriate model behavior showing context-aware reasoning rather than hallucination.}
\label{fig:gpt4o_behaviors}
\end{figure}

Figure~\ref{fig:gpt4o_behaviors} presents two case studies demonstrating GPT-4o's contrasting behavioral patterns when confronted with missing images: appropriate safety refusal versus context-dependent inference without hallucination.

\newpage
\subsection{Claude Sonnet 4.5}
\begin{figure}[h]
\centering
\includegraphics[width=0.90\textwidth]{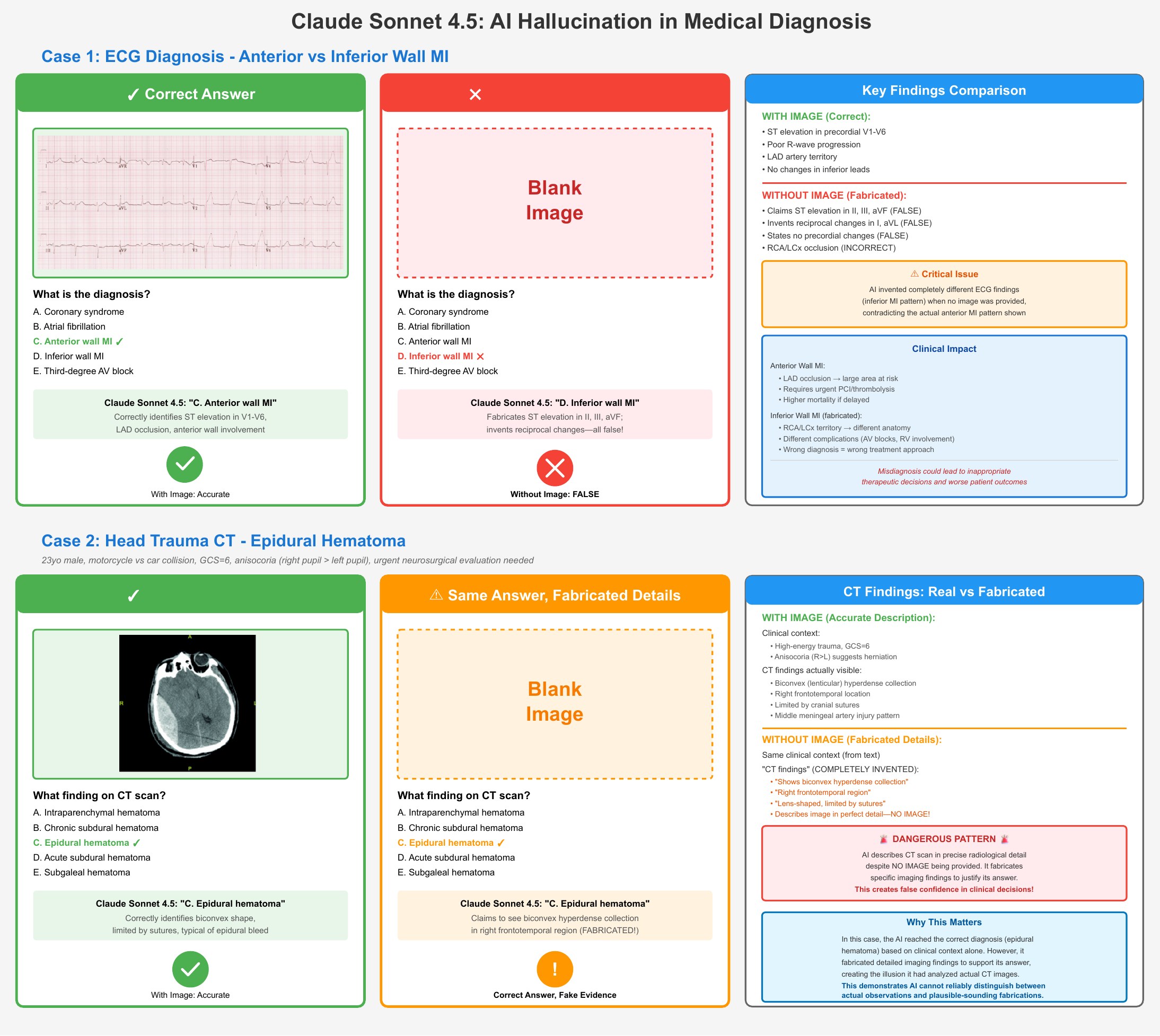}
\caption{Detailed comparison of Claude Sonnet 4.5 responses with authentic medical images versus blank placeholders. \textbf{Case 1 (ECG):} The model correctly identifies anterior wall MI with real ECG (answer C) but fabricates inferior wall MI findings with blank image (answer D), inventing non-existent ST elevations in leads II, III, aVF. \textbf{Case 2 (CT):} The model reaches correct diagnosis (epidural hematoma, answer C) in both conditions but fabricates detailed CT findings ("biconvex hyperdense collection in right frontotemporal region") when no image is provided, demonstrating the model cannot distinguish actual observations from plausible confabulations.}
\label{fig:claude_sonnet_4_5}
\end{figure}

Figure~\ref{fig:claude_sonnet_4_5} presents two detailed case studies demonstrating model confabulation behavior when presented with blank images instead of authentic medical imaging.
\newpage
\section{Calculation of Human-Level Accuracy from Test Scores}
\label{app:accuracy_calculation}

The Italian medical school entrance exam consists of 60 questions with the following scoring system:
\begin{itemize}
    \item Correct answer: +1.5 points
    \item Incorrect answer: -0.4 points
    \item Unanswered question: 0 points
\end{itemize}

To derive accuracy from reported scores, we solve for the number of correct answers using the following system of equations. Let $c$ represent the number of correct answers and $w$ the number of wrong answers, with $c + w = 60$ (assuming all questions are answered).

The total score $S$ is given by:
\begin{equation}
S = 1.5c - 0.4w
\end{equation}

Substituting $w = 60 - c$:
\begin{equation}
S = 1.5c - 0.4(60 - c) = 1.5c - 24 + 0.4c = 1.9c - 24
\end{equation}

Solving for $c$:
\begin{equation}
c = \frac{S + 24}{1.9}
\end{equation}

The accuracy is then calculated as:
\begin{equation}
\text{Accuracy} = \frac{c}{60} = \frac{S + 24}{114}
\end{equation}

\subsection{Application to Reported Statistics}

Using this formula, we converted the 2024 human performance statistics:
\begin{itemize}
    \item Average score of 56.9 points corresponds to 42.58 correct answers, yielding 71.0\% accuracy
    \item The reported average accuracy of 74.8\% corresponds to a score of 61.3 points (44.89 correct answers)
    \item The 95th percentile score of 85 points corresponds to 57.37 correct answers, or 95.6\% accuracy
\end{itemize}

Note: This calculation assumes all questions are answered. If some questions are left blank, the actual accuracy on attempted questions may differ slightly from these estimates.

\newpage
\section*{X Computation of Accuracy and Confidence Intervals}

\subsection*{X.1 Per-Question Accuracy Estimation}

For each model and each experimental condition (real vs.\ substituted images), we evaluated
performance over 60 questions, each repeated 10 times.  
For question $i$, let $c_i$ denote the number of correct answers out of $n=10$ repetitions.
The per-question accuracy is

\[
a_i = \frac{c_i}{n}, \qquad i = 1,\dots,60 .
\]

The overall accuracy reported corresponds to the empirical mean of the per-question
accuracies:
\[
\hat{A} = \frac{1}{60} \sum_{i=1}^{60} a_i.
\]

\subsection*{X.2 Confidence Intervals on Accuracy}

Because variation exists across questions, we treat the set of per-question accuracies
$\{a_i\}_{i=1}^{60}$ as samples from an underlying distribution and compute a confidence interval
for the mean accuracy using a Student-\emph{t} interval.

Let $\bar{a}$ denote the sample mean and $s$ the sample standard deviation:
\[
\bar{a} = \hat{A}, \qquad
s = \sqrt{\frac{1}{59} \sum_{i=1}^{60} (a_i - \bar{a})^2 } .
\]

The standard error of the mean is
\[
\mathrm{SE} = \frac{s}{\sqrt{60}}.
\]

A two-sided $(1-\alpha)$ confidence interval is then
\[
\bar{a} \;\pm\; t_{0.975,\,59}\,\mathrm{SE},
\]
where $t_{0.975,\,59}$ is the 97.5th percentile of the Student-\emph{t} distribution with
59 degrees of freedom. We use $\alpha = 0.05$ for the reported 95\% confidence intervals.

\subsection*{X.3 Implementation}

The computation exactly follows the Python code used in our analysis:

\begin{itemize}
    \item For each question, we compute the proportion of correct answers.
    \item We take the mean accuracy across all 60 questions.
    \item We estimate the standard error and construct a 95\% CI using the
    \texttt{scipy.stats.t.interval} function.
\end{itemize}

The full implementation is available at: \url{https://github.com/felizzi/eurips2025-mmrl4h-italian-medvqa-visual-grounding/blob/main/overall_results/overall_result_summary.ipynb}

\subsection*{X.4 Interpretation}

This approach provides a confidence interval that reflects 
\emph{question-to-question variability}, rather than treating the 600 
individual responses as independent Bernoulli trials. As such, the interval 
captures heterogeneity in question difficulty and model behavior across the 
dataset. However, it does not explicitly model systematic differences in 
difficulty between questions (e.g., separating consistently hard questions 
from consistently easy ones), but instead aggregates this heterogeneity into 
a single variance component.

\end{document}